\documentclass[conference, letterpaper]{IEEEtran}
\usepackage{cite}
\usepackage{algorithm}
\usepackage{algpseudocode}
\usepackage{subcaption}
\usepackage{ragged2e}

\usepackage{flushend} 

\ifCLASSINFOpdf
   \usepackage[pdftex]{graphicx}
\else
\fi

\usepackage[cmex10]{amsmath}
\usepackage{color}
\usepackage{fancyhdr}

\renewcommand{\thispagestyle}[2]{}

\fancypagestyle{plain}{
        \fancyhead{}
        \fancyhead[C]{first page center header}
        \fancyfoot{}
        \fancyfoot[C]{first page center footer}
}
\begin{document}

%
\title{KNN and ANN-based Recognition of Handwritten Pashto Letters using Zoning Features}

\author{\IEEEauthorblockN{Sulaiman Khan$^1$,Hazrat Ali$^2$, Zahid Ullah$^3$, Nasru Minallah$^4$, Shahid Maqsood$^5$, and Abdul Hafeez$^6$}\\

\IEEEauthorblockA{Dept. of CS, University of Swabi, Pakistan$^1$,\\ Department of Electrical and Computer Engineering, COMSATS University Islamabad, Abbottabad Campus, Pakistan$^2$\\
Dept. of EE, CECOS University, Pakistan$^3$, Dept. of CS, UET, Jalozai Campus, Pakistan$^{4,6}$, Dept. of IE, UET, Pakistan$^5$\\
Correspondence: hazratali@cuiatd.edu.pk, zahidullah@cecos.edu.pk, abdul.hafeez@uetpeshawar.edu.pk
}}
\maketitle

\begin{abstract}
This paper presents a recognition system for handwritten Pashto letters. However, handwritten character recognition is a challenging task. These letters not only differ in shape and style but also vary among individuals. The recognition becomes further daunting due to the lack of standard datasets for inscribed Pashto letters. In this work, we have designed a database of moderate size, which encompasses a total of 4488 images, stemming from 102 distinguishing samples for each of the 44 letters in Pashto. The recognition framework uses zoning feature extractor followed by K-Nearest Neighbour (KNN) and Neural Network (NN) classifiers for classifying individual letter. Based on the evaluation of the proposed system, an overall classification accuracy of approximately 70.05\% is achieved by using KNN while 72\% is achieved by using NN.
\end{abstract}


\IEEEpeerreviewmaketitle

section{Introduction}
\label{sec:Introduction}
In this modern technological and digital age, optical character recognition (OCR) systems play a vital role in machine learning and automatic recognition problems. OCR is a section of software tool that converts printed text and images to machine readable form and enables the machine to recognize images or text like humans. 
OCR systems are commercially available for isolated languages, which include Chinese, English, Japanese, and others. However, few OCR systems are available for cursive languages such as Persian and Arabic and are not highly robust. To the best of our knowledge, there is no such commercial OCR system available for carved Pashto letters recognition; however, such systems exist in research labs. 

Handwritten letters recognition is a daunting task mainly because of variations in writing styles of different users. Handwritten letters recognition can be done either offline or online. Online character recognition is simpler and easier to implement due to the temporal based information such as velocity, time, number of strokes, and direction for writing. In addition, the trace of the pen is a few pixels wide so this does not require thinning techniques for classification. On the other hand, offline character recognition system implementation is even laborious due to high variations in writing and font styles of every user. In our paper, we present inscribed of handwritten Pashto letters.

Pashto is a major language of Pashtun tribe in Pakistan and the official language of Afghanistan. In censes 2007 – 2009, it was estimated that about 40 – 60 millions of people around the world are native speakers of this language.
Pashto letters can be shaped into six different formats, which make the recognition process challenging. Furthermore, the count of character dots and occurrence of these dots that varies from letter to letter make the problem challenging. 
In order to address these problems, research shows the use of high level features based on the structural information of letters. An OCR based system  using deep learning network model that incorporates Bi- and Multi-dimensional long short term memory for printed Pashto text recognition has been suggested~\cite{ahmad2015recognizable}. 

A web-based survey shows that Pashto script contains a huge number of unique ligature~\cite{ahmad2016kpti}.
Such ligature makes the implementation of OCR system for carved Pashto challenging. As printed letters contain a constant shape/style and font size; thus, the said technique fails in our case due to higher higher variations in style and font in case of inscribed letters. Riaz et al.\cite{ahmad2015robust} has presented the development of an OCR system for cursive Pashto script using scale invariant feature transform and principle component analysis. In order to address this issue, we present a system for handwritten Pashto letters recognition, which has the following key contributions:

\begin{itemize}
 \item As there is no standard handwritten Pashto letters database for testing an algorithm; thus, one of the  contribution of this work is to develop and present a medium-sized database of 4488 (102 samples for each letter) for further research work.
 \item The second contribution of this research work is to provide a base result as a benchmark for Pashto language. For this purpose, the performance results of the state-of-the-art classifiers$-$KNN and deep Neural Network are used based on zoning features. 
 \item Our proposed handwritten Pashto letters recognition system is efficient, simple, and cost-effective.
 \item We provide comprehensive results for analyzing the proposed system for handwritten Pashto letters recognition, which may help the researchers to further explore this area.
 \end{itemize} 

This paper is divided in seven sections: Section~\ref{sec:relatedwork} explains the related work. Section~\ref{sec:backgroundstudy} captures the background information about the classifiers and feature extraction algorithm used in this research work. Section~\ref{sec:The Proposed Methodology} delineates the methodology. Section~\ref{sec:Feature Extraction} discusses about the feature extraction, which is very important in the area of pattern recognition and machine learning while section~\ref{sec:results} demonstrates the experimental results followed by the conclusions and future work in Section~\ref{sec:Conclusions and Future work}.

\section{Related Work}
\label{sec:relatedwork}
Pashto, Persian, Urdu, and Arabic are sister languages. Several diverse approaches are suggested by different researchers for developing an OCR system for these languages. However, Pashto script contains more letters (44) than Arabic script (28 letters), Persian script (32 letters), and Urdu script (38 letters). Pashto language encapsulates all the letters from Urdu script with additional seven letters. This additional seven letters make the OCRs developed for Persian, Urdu, and Arabic language unable to recognize handwritten Pashto letters. As per our best knowledge, some of the closely related work on the prescribed languages is mentioned below.

Abdullah et al. \cite{abdulllah2018arabic} presented an OCR system for Arabic handwriting recognition based on Neural Network classifier for classifying an IFN|ENIT dataset. Ahmad et al. \cite{ahmad2017ligature} presented a novel approach of gated bidirectional long short term memory (GBLSTM) for recognition of printed Urdu Nasta’liq text, which is a special form of Neural Network based on ligature information of the printed text. 
Ahmed et al. \cite{ahmed2017handwritten} used a one dimensional BLSTM for handwritten Urdu letter recognition where a medium size database for handwritten Urdu letters collected from 500 people was developed.

Alotaibi et al. \cite{alotaibi2018optical} suggested an algorithm to develop an OCR that can check the originality and similarity of online Quranic contents where Quranic text is a combination of diacritics and letters. For diacritic detection, they used region-based algorithms and projection method is used for letter detection. The results of the similarity indices are compared with standard Mushaf Al Madina benchmark.
Boufenar et al. \cite{boufenar2018artificial} presented the concept of supervised learning technique named Artificial immune system based on zoning technique for isolated carved Arabic letters recognition.
Jameel and Kumar \cite{jameel2017offline} suggested the use of B spline curves as a feature extractor for offline Urdu character recognition. Naz et al. \cite{naz2017urdu}\cite{naz2018pak} presented the use of multi-dimensional recurrent Neural Network based on statistical features for Urdu Nasta’liq text recognition.
Rabi et al. \cite{rabi2017survey} performed a survey on different OCR systems for handwritten cursive Arabic and Latin script recognition where it was concluded that the results of contextual sub character of Hidden Markov Models were proven with high accuracy for handwritten Latin and Arabic script recognition.

Rouini et al. \cite{rouini2017off} presented the use of dynamic random forest classifier based on surf descriptor feature extraction technique. Sahlol et al. \cite{sahlol2017arabic} inspected different classifiers Genetic algorithm (GA), Particl Swam optimization (PSO), Grey Wolf optimization (GWO), and BAT algorithms (BAT) for handwritten Arabic characters recognition. After testing each algorithm, it was concluded that GWO provides prominent results for handwritten Arabic characters recognition. As Sindhi language is a super set of Arabic language, Shaikh et al. \cite{shaikh2008segmentation} developed an OCR system for text recognition using an approach based on segmentation.

M. Kumar et al. \cite{kumar2018character} presented a comprehensive survey of Indic and non-Indic scripts on letters and numeral recognition. Zayene et al. \cite{zayene2018multi} presented a novel approach for Arabic video text recognition using recurrent Neural network. This system suggests a segmentation free method mainly based on a multi-dimensional version of long short term memory combined with a connectionist temporal classification layer.
Veershetty et al. \cite{veershetty2018radon} suggested the concept of an optical character recognition (OCR) system for handwritten script recognition based on KNN, SVM, and linear discriminant analysis (LDA) classifiers. For feature extraction, they used a technique based on Radon and wavelet transform, and words were extracted using morphological dilation methods.

Malviya et al. \cite{malviya2018feature} carried out a comparative study of various feature extractions techniques named Zernike moments, projection histogram, zoning methods, template machine, and chain coding technique and classification algorithms such as SVM and Artificial Neural Network (ANN) have been discussed. Some vital parameters are selected based on sample size, data types, and accuracy. 
Bhunia et al.\cite{bhunia2018indic} presented a novel approach for word level Indic-script recognition using character level data in input stage. This approach uses a multimodal Neural Network that accepts both offline and online data as an input to explore the information of both online and offline modality for text/script recognition. This multi-modal fusion scheme combines the data of both offline and online data, which indeed a real scenario of data being fed to the network. The validity of this system was tested for English and six Indian scripts. 
Obaidullah et al. \cite{obaidullah2018handwritten} carried out a comprehensive survey for the development of an OCR system for Indic script recognition in multi-script document images. Multiple pre-processing techniques, feature extraction techniques, and classifiers used in script recognition were discussed.

The literature review shows that a little work is available on the development of an OCR system for the recognition of printed Pashto letters; however, there is no OCR system developed for automatic recognition of handwritten Pashto letters. All the above mentioned algorithms perform well for the specified languages but fail in recognizing the handwritten Pashto letters owing to the extra number of letters in the character set. In this paper, we present a robust OCR system for the recognition of handwritten Pashto letters having the key benefits mentioned above.

\section{Background Study}
\label{sec:backgroundstudy}
This part of the paper describes the background detail of the character modeling for Pashto script, classification techniques followed by KNN, and Neural Network classifiers.

\subsection{Pashto}
\label{sec:pashto}
Pashto is the language of Pashtuns, often pronounced as Pakhto/Pukhto/Pushto and is the official language of Afghanistan and a major language of Pashtun clan in Pakistan.  In Persian literature, it is known as Afghani while in Urdu or Hindi literature, it is known as Pathani. Pashto has two major dialects namely “soft dialect” and “hard dialect”. Both of these dialects are phonologically differ from each other. The soft dialect is called “southern” while the hard dialect is known as “northern”. In soft dialect i-e., “southern”, Pushto is spelled as Pashto while in hard dialect i-e., in “northern”, it is spelled as Pukhto or Pakhto. The word Pashto is followed as a representation for both hard and soft dialects. The Kandahari form of Pashto dialect, also known as “Pata Khazana”, is considered as standard spelling system for Pashto script.

\begin{figure}[!t]
  \centering
  \includegraphics[width=2.7in]{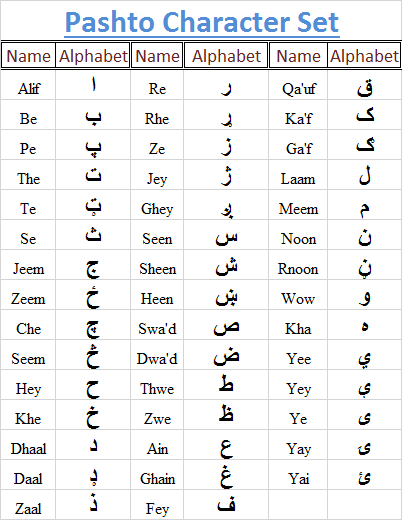}\\
  \caption{Pashto characters dataset.}\label{fig_pashto}
\end{figure}

Pashto script consists of 44 letters shown in Fig.~\ref{fig_pashto}. The “Name” represents letter name while “Alphabet” represents letters shape in isolated form. It has borrowed all the letters from Persian script, i.e., 32 letters that has further borrowed the entire letter set, i.e., 28 letters from Arabic script. That is why Pashto is known as a modified pattern of Perso-Arabic characters. Urdu script adopts all 32 letters from Persian script with 6 additional letters. Pashto script encapsulates all the Urdu characters with minor change in these 6 special characters for Urdu script as shown in Table~\ref{tbl:urduSpecific}. It encompasses additional 7 characters, especially to Pashto script forming a dataset of 44 characters as shown in Table~\ref{tbl:pashtoSpecial}. In order to make a word in Pashto script, two or more than two isolated letters are combined to form a word. While defining a word, a letter shape changes w.r.t its position (start, middle or end) in the word as shown in the Table~\ref{tbl:letterShape}. Both Naksh and Nasta’liq is followed for Pashto script writing; however, Naksh is considered as standard writing style for Pashto script.

\begin{table}[!t]
  \centering
    \begin{minipage}{.3\textwidth}
      \includegraphics[width=\linewidth]{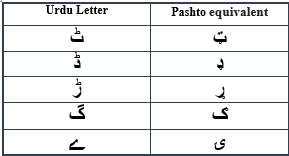}\\
    \end{minipage}
  \caption{Urdu specific letters representation in Pashto script.}\label{tbl:urduSpecific}
\end{table}

\begin{table}[!t]
  \centering
    \begin{minipage}{.3\textwidth}
      \includegraphics[width=\linewidth]{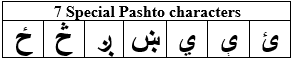}\\
    \end{minipage}
  \caption{Pashto specific letters.}\label{tbl:pashtoSpecial}
\end{table}

\begin{table}[!t]
  \centering
    \begin{minipage}{.3\textwidth}
      \includegraphics[width=\linewidth]{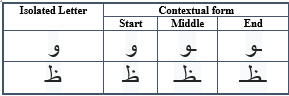}\\
    \end{minipage}
  \caption{Change in letter’s shape w.r.t. its position in word.}\label{tbl:letterShape}
\end{table}

\subsection{K-Nearest Neighbor (KNN)}
KNN is a supervised learning tool used in regression and classification problems. In training phase, KNN uses multi-dimensional feature vector space that assigns a class label to each training sample. Many researchers have suggested the use of KNN classifier in text/digits recognition and classification such as Hazra et al~\cite{hazra2017optical} who presented the concept of KNN classifier for  both handwritten and printed letters recognition in English language based on sophisticated feature extractor technique. 

For online handwritten, Gujarati character recognition Naik et al. \cite{naik2017online} suggested the use of SVM with polynomial, linear, and RBF kernel, KNN with variant values of K and multi-layer perception (MLP’s) for stroke classification based on hybrid feature set. Selamat et al. \cite{selamat2009arabic} suggested the use of hybrid KNN algorithms for web paged base Arabic language identification and classification. They carried out the results based on SVM, back propagation neural network, KNN, and hybrid KNN.
Zhang et al. \cite{zhang2006svm} presented the use of KNN for visual category recognition based on text, color, and particularly shape in a homogeneous framework. Hasan \cite{hassan2018arabic} presented the concept of KNN classifier for Arabic(Indian) digits recognition using multi-dimensional features, which consist of discrete cosine transform (DCT) and projection methods.

\begin{figure}[!t]
  \centering
  \includegraphics[width=2.5in]{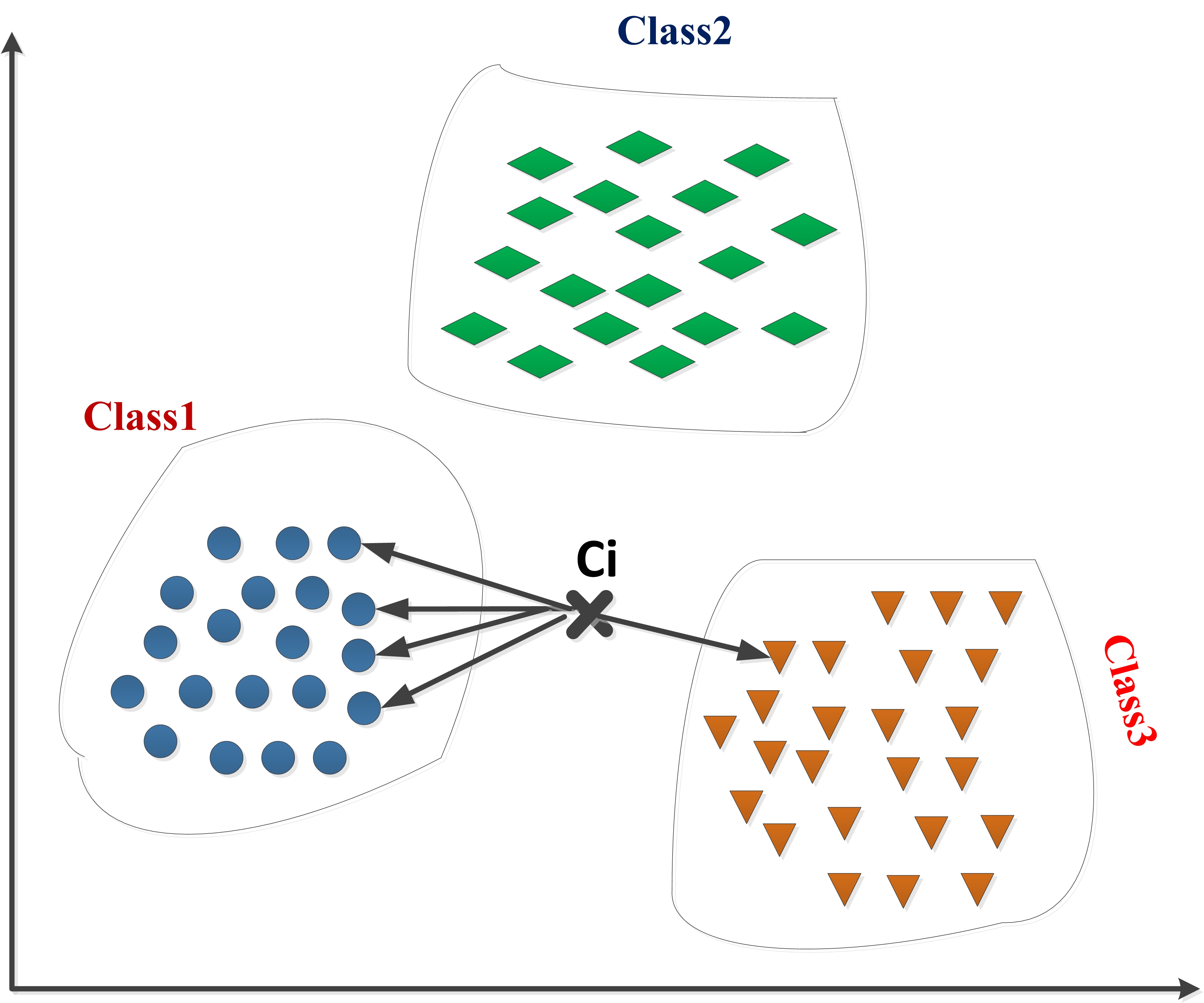}\\
  \caption{Basic multi-class KNN basic model.}\label{fig:KNN}
\end{figure}

KNN generates classification results by storing all the available cases and stratify new classes based on a similarity measure (distance functions). Pashto contains 44 letters in its character set so there are 44 classes to be classified. In short, it is a multi-class recognition problem. Fig.~\ref{fig:KNN} represents a basic multi-class KNN model.
In Fig.~\ref{fig:KNN} class1, class2, and class3 represent 3 different classes. In our case, it contains 44 classes as there are 44 letters in Pashto character dataset.

\subsection{Neural Network (NN)}
NN has performed a vital role in the recognition and classification problems. Inspired from human nervous system, ANN is composed of layered architecture\textemdash input, hidden, and output layer. It contains a network of neurons connected through weighted connections that accepts input, performs processing, and produce detailed patterns. Machine learning (ML) has been widely used in a varitey of applications. ML has been used in scheduling tasks in real time through cloud computing in the form of genetic algorithms \cite{mahmood2017hard}.
Another study shows the use of ML models in genomics \cite{krachunov2017application}. The goal is to detect variations and errors in Genomics datasets that entail higher variations. Decision trees and tabu search have been utilized in order to learn the dispatching rules for smart scheduling \cite{shahzad2016learning}. To explore the active learning, exponential gradient exploration has been studied \cite{bouneffouf2016exponentiated}.

\begin{figure}[!t]
  \centering
  \includegraphics[width=2.5in]{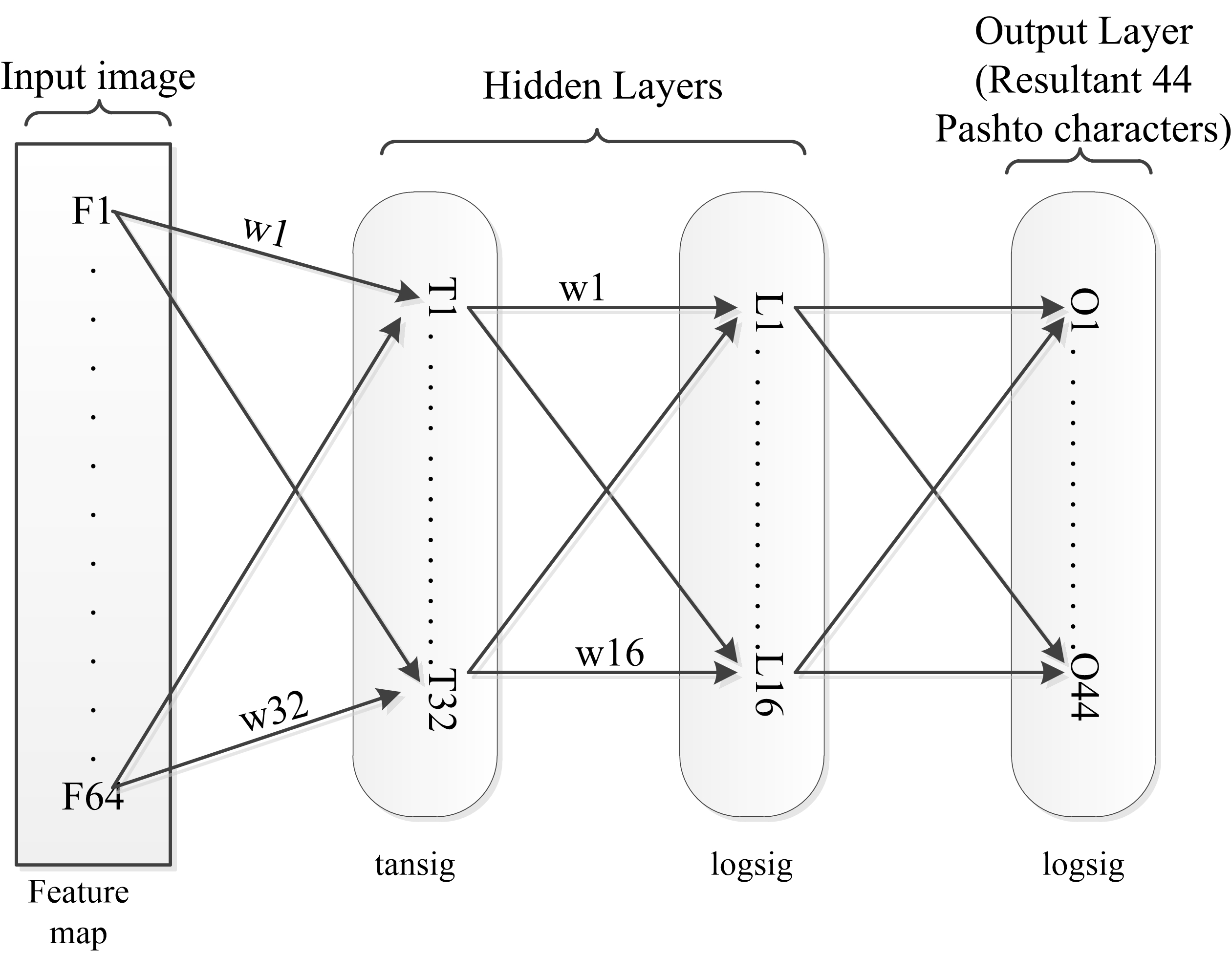}\\
  \caption{Neural Network for handwritten Pashto character recognition.}\label{fig:ANN}
\end{figure}

Owing to NN's high identification and recognition abilities especially in text recognition problems, multiple researchers have suggested the use of this model, some of which are mentioned here.
Jameel et al. \cite{jameel2017review} carried a review paper on Urdu character recognition using NN. In this paper, they suggested the use of B-Spline curves as a feature extractor technique for Urdu characters recognition. Zhang et al. \cite{zhang2018drawing} presented the use of recurrent NN for drawing and recognition purposes of Chinese language.
Patel et al. \cite{patel2017wavelet} suggested the use of ANN for handwritten character recognition based on discrete wavelet transform as a feature extractor technique, which is based on accurate level of multi-resolution technique. A basis NN diagram for HPLR system is shown in Fig.~\ref{fig:ANN}. In this research work, a NN classifier is selected with two hidden layers and one input and output layer. A feature map of 16 distinct values based on zoning technique are fed at input layer and the expected results are calculated at the output layer.

\section{The Proposed Methodology}
\label{sec:The Proposed Methodology}
\begin{figure}[!t]
  \centering
  \includegraphics[width=2.5in]{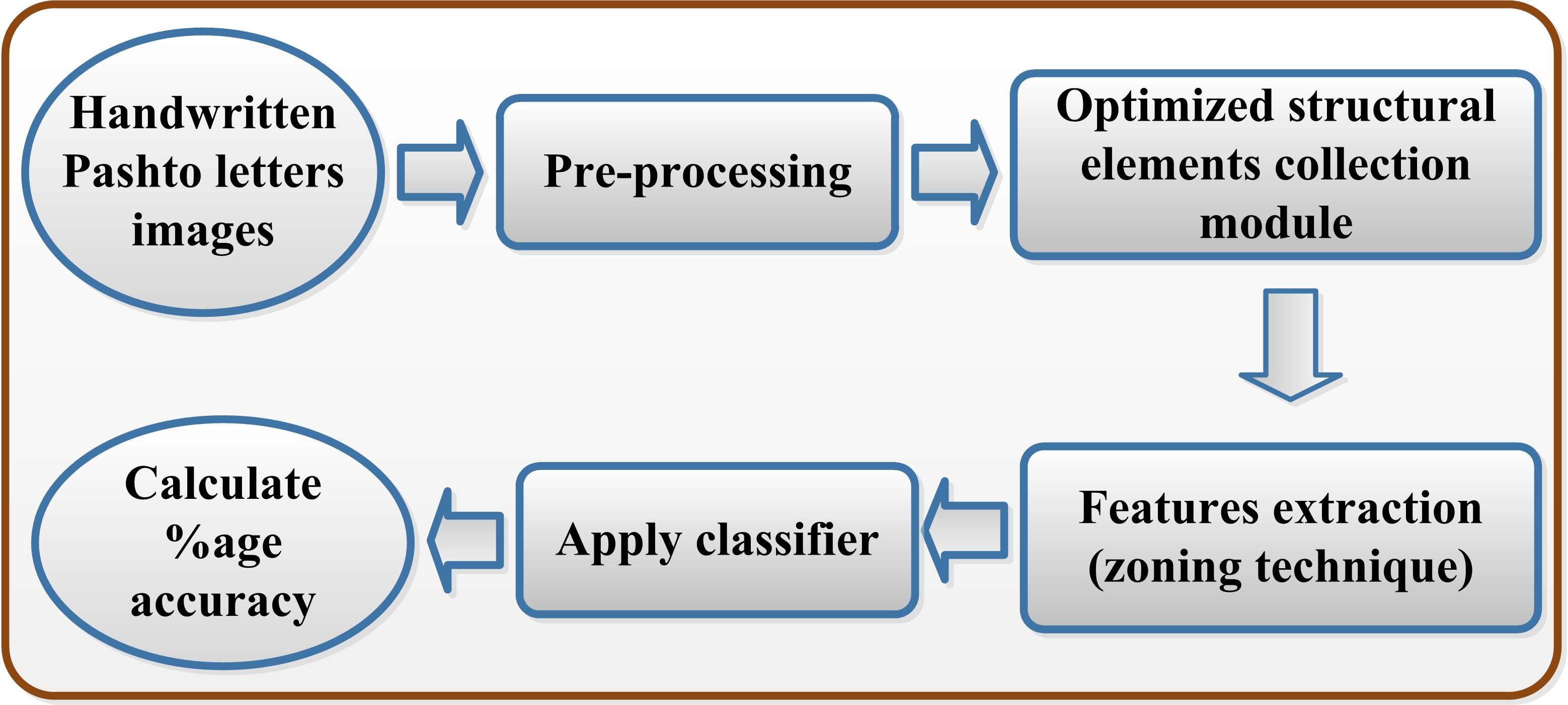}\\
  \caption{The proposed Pahsto handwritten letter recognition system.}\label{fig:BlockDiagram}
\end{figure}

The proposed OCR system for the recognition of handwritten Pashto letters is divided into three main steps as shown in Fig.~\ref{fig:BlockDiagram}.

\begin{itemize}
  \item Database development for the handwritten Pashto letters.
  \item Feature selection/extraction.
  \item Classification and recognition using KNN and NN classifiers.
\end{itemize}

\subsection{Database development for the handwritten Pashto letters}
\begin{figure}[!t]
  \centering
  \includegraphics[width=2.5in]{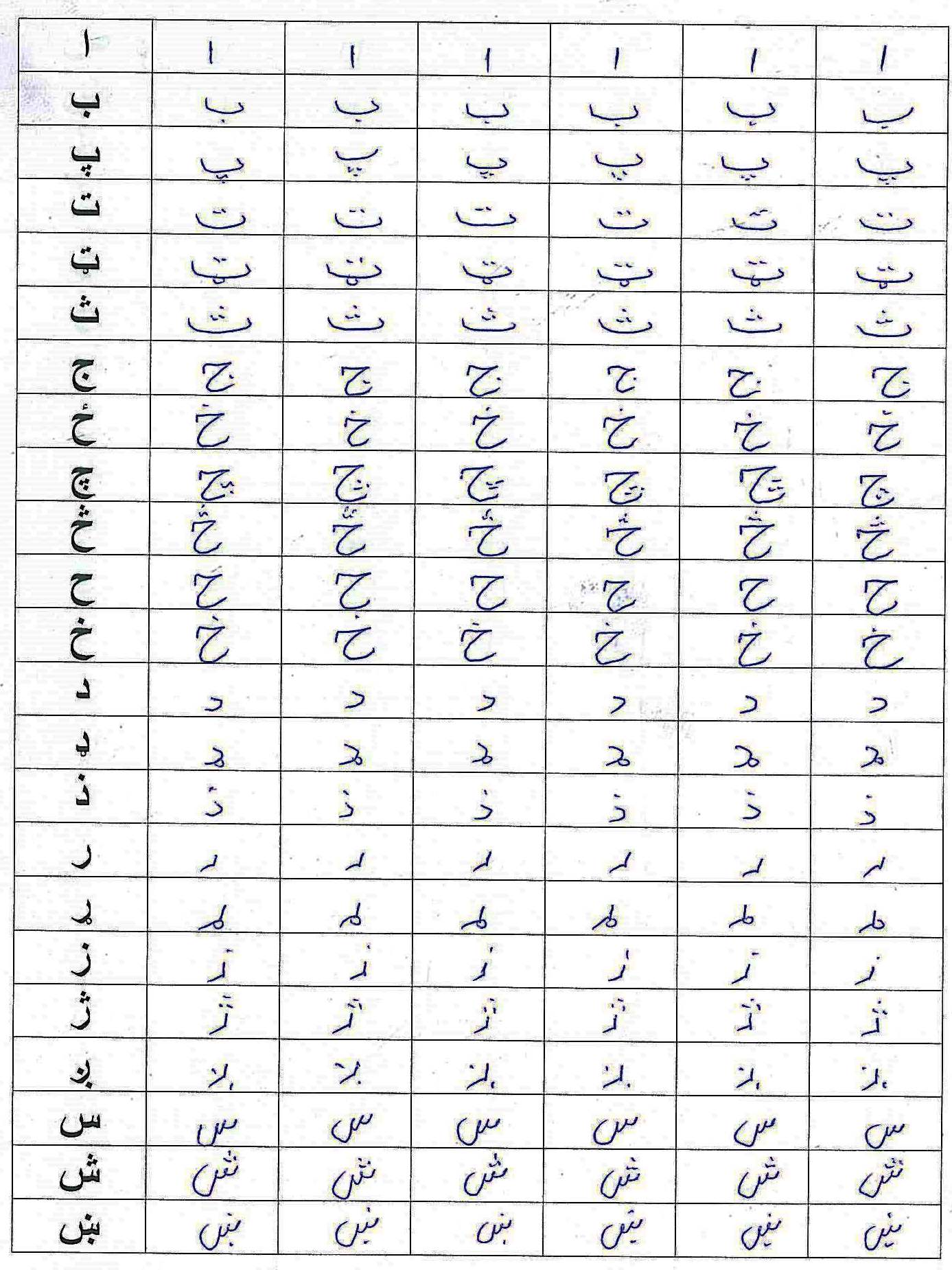}\\
  \caption{First 23 handwritten Pashto characters.}\label{fig:first23}
\end{figure}

\begin{figure}[!t]
  \centering
  \includegraphics[width=2.5in]{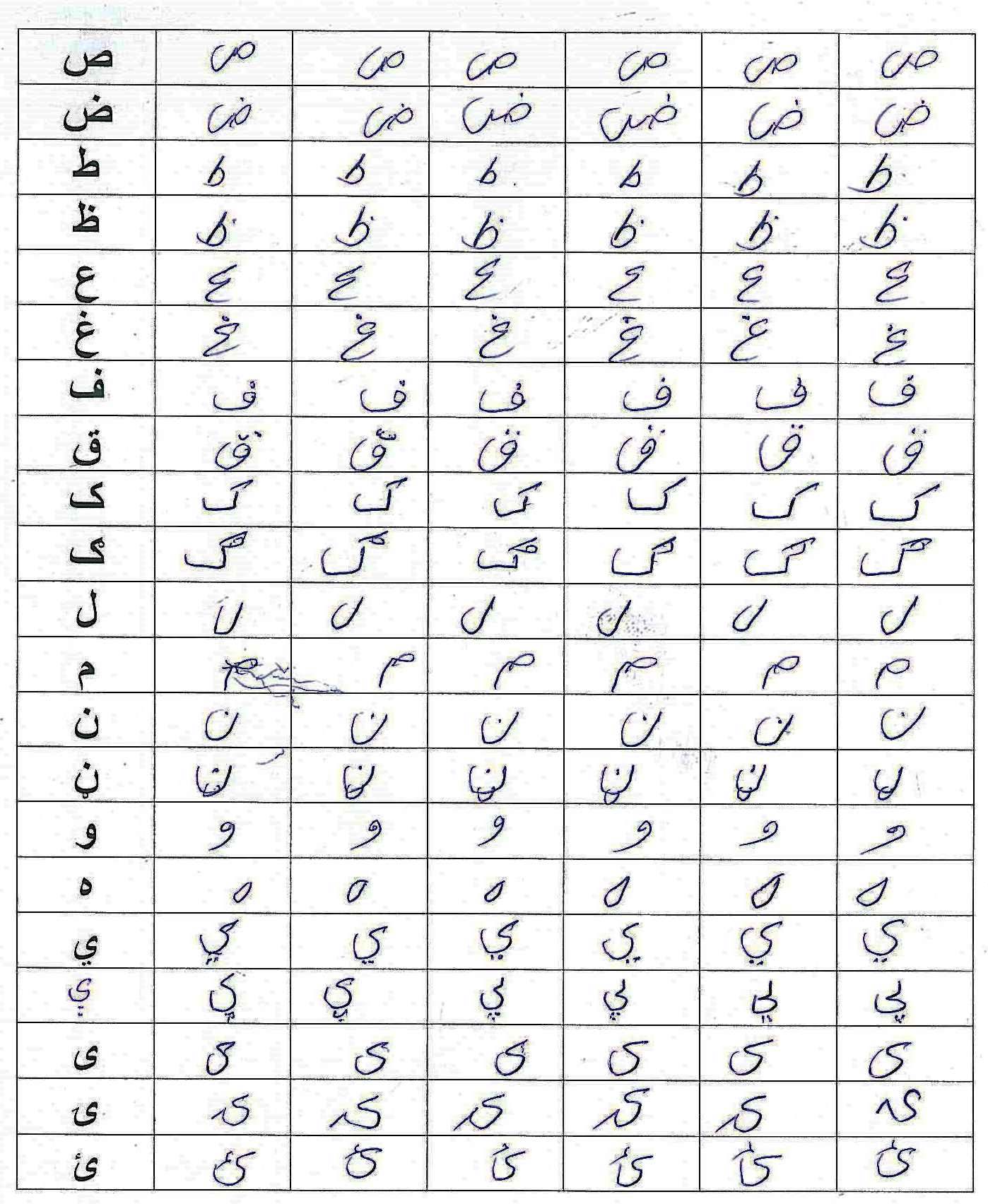}\\
  \caption{Remaining 21 handwritten Pashto characters.}\label{fig:remaining21}
\end{figure}

A medium size handwritten character database of 4488 characters (contains 102 samples for each letter) is developed by collecting handwritten samples from different individuals. These samples are collected on an A4 size paper divided into 6 columns for collecting a letter variant samples from same person. These samples are further scanned into computer readable format as shown in as shown in Fig.~\ref{fig:first23} and Fig.~\ref{fig:remaining21}.

\subsubsection{Letters Extraction}
The letters are extracted in order to create a database. A few extracted letters are shown in Table~\ref{tab:sliced}. All these extracted letters are resized into a fixed size of 44$\times$44. This fixed size of the character helps in generating a uniform sized feature vector.

\begin{table}[!t]
\caption{A table with handwritten sliced Pashto characters.}
\centering
\begin{tabular}{cc}
\begin{subfigure}{0.2\textwidth}\centering\includegraphics[width=0.3\columnwidth]{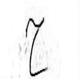}\caption{}\end{subfigure}&
\begin{subfigure}{0.2\textwidth}\centering\includegraphics[width=0.3\columnwidth]{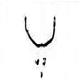}\caption{}\end{subfigure}\\
\newline
\begin{subfigure}{0.2\textwidth}\centering\includegraphics[width=0.3\columnwidth]{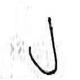}\caption{}\end{subfigure}&
\begin{subfigure}{0.2\textwidth}\centering\includegraphics[width=0.3\columnwidth]{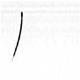}\caption{}\end{subfigure}\\
\end{tabular}
\label{tab:sliced}
\end{table}

Each extracted letter in Table~\ref{tab:sliced} is hugely affected with dark spots i.e., noise, which is removed using thresholding. 
During the data collection phase, the handwritten character position varies in the 64$\times$64 region/box. The reason is that letters can be written on top, left, right, and bottom of the box varying from person to person. We have centralized all the letters. Post-thresholding and centralizing results are captured in Table~\ref{tab:thresholded}.

\begin{table}[!t]
\caption{A table with thresholded and centralized results on sliced characters.}
\centering
\begin{tabular}{@{\hskip2pt}c@{\hskip2pt}c@{\hskip2pt}c}
\begin{subfigure}{0.2\textwidth}\centering\includegraphics[width=0.3\columnwidth]{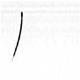}\caption{Befor thresholding}\end{subfigure}&
\begin{subfigure}{0.2\textwidth}\centering\includegraphics[width=0.3\columnwidth]{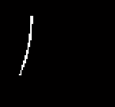}\caption{After thresholding}\end{subfigure}\\
\begin{subfigure}{0.2\textwidth}\centering\includegraphics[width=0.3\columnwidth]{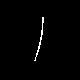}\caption{Centralized}\end{subfigure}
\end{tabular}
\label{tab:thresholded}
\end{table}

\section{Feature Extraction}
\label{sec:Feature Extraction}
Selecting an astute, informative and independent feature is a crucial step for effective classification. This paper presents the concept of zoning method as a feature extractor technique for the recognition of handwritten Pashto letters.

\subsection{Zoning Technique}
\begin{figure}[!t]
  \centering
  \includegraphics[width=2.5in]{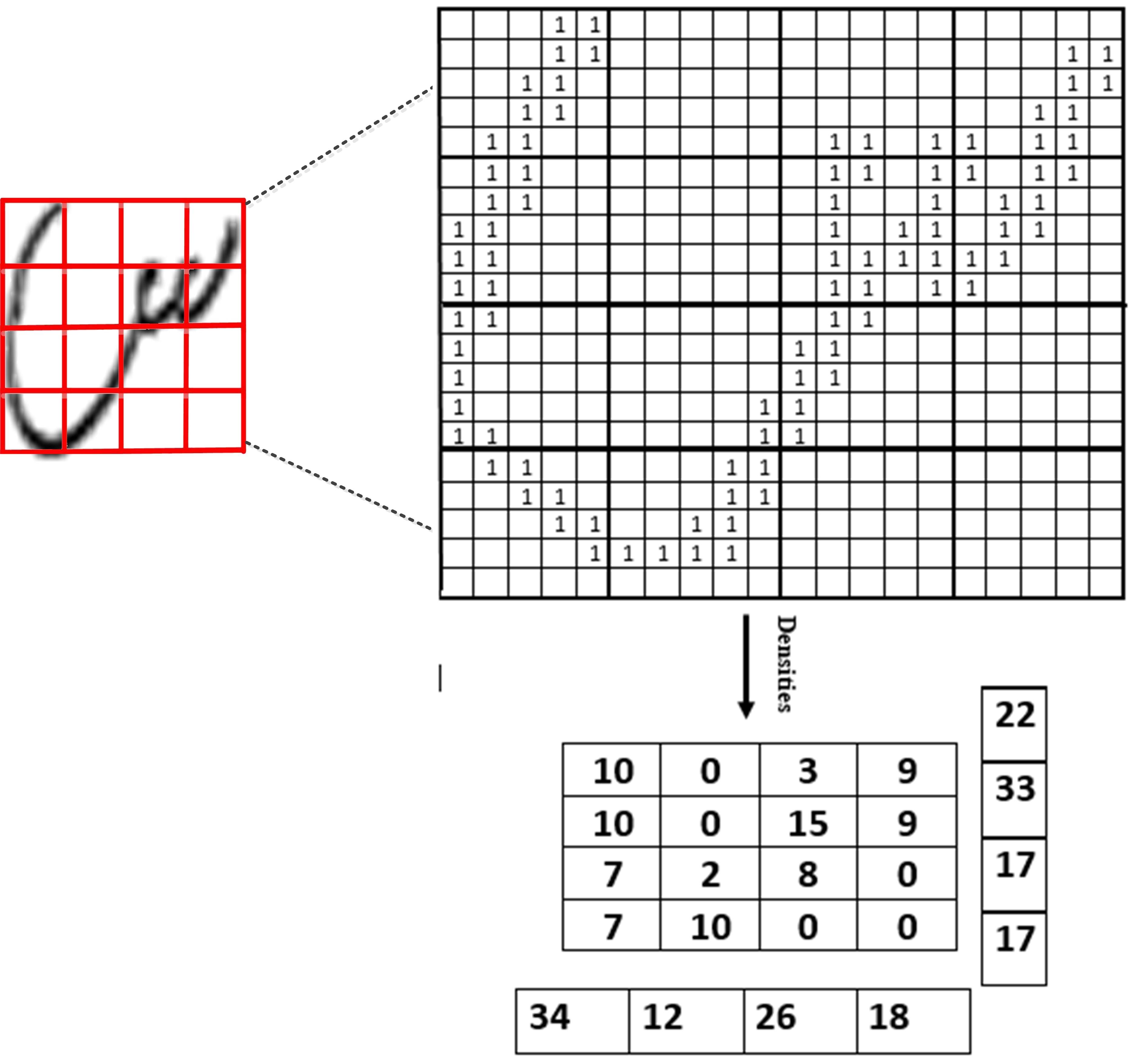}\\
  \caption{Zoning feature extraction.}\label{fig:zoning}
\end{figure}

This research work uses a 4$\times$4 static grid to extract each letter features as shown in Fig.~\ref{fig:zoning}. By applying this zoning grid, it superimposes the pattern/character image and divides it into 16 equal zones. In each zone, the density of the letter is extracted that represents the ratio of the black pixels forming the letter on the total size of zone \cite{nebti2013handwritten}. In this way, a feature map for all 4488 letters is obtained for classification.
After applying this technique a feature vector of 16 real values formed for each sample because we focuses on zones not on the number of pixels.

\section{Results}
\label{sec:results}
This section summarizes the results obtained after applying KNN and NN classifiers to handwritten Pashto letters for classification/recognition.

\subsection{Classification Accuracy of K - Nearest Neighbours}
\begin{figure*}[ht]
\centering
\includegraphics[width=5.0in]{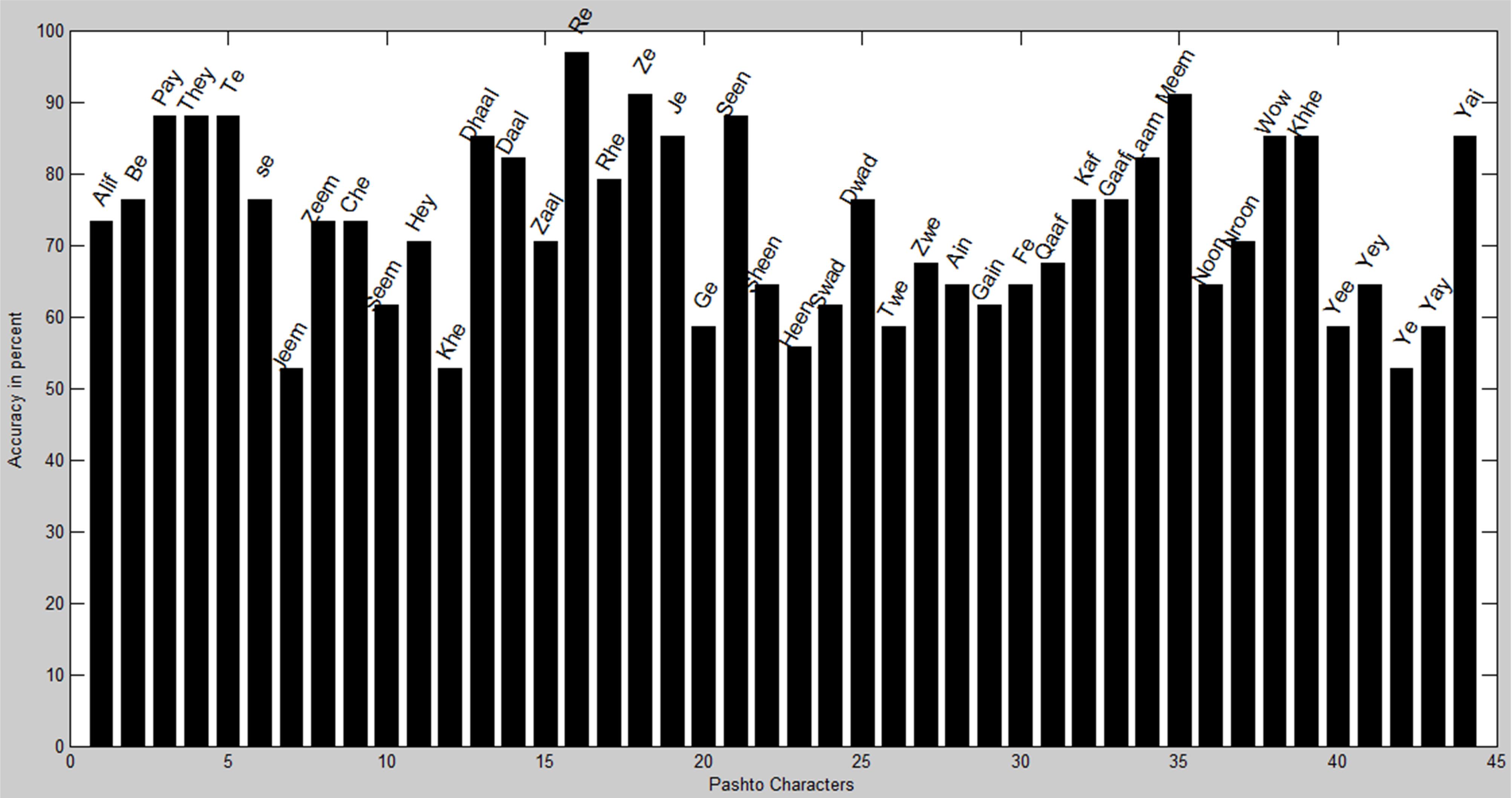}\\
\caption{KNN classifier accuracy results for HPLR system.}\label{fig:resultKNN}
\end{figure*}

The results of the KNN classifier for Pashto script recognition are shown in Fig.~\ref{fig:resultKNN}. The results are carried out using KNN classifier based on zoning features. The total image features for the Pashto letters is divided into a ratio of (2:1) for training and testing phases. The databases consists 102 samples for each Pashto letter. Thus, 68 letters features are selected for training phase and the remaining 34 letters features are selected for testing phase. An overall accuracy of about 70.05\% is obtained for KNN, lesser than ANN, which is 72\%.

\begin{figure}[!t]
  \centering
  \includegraphics[width=2.5in]{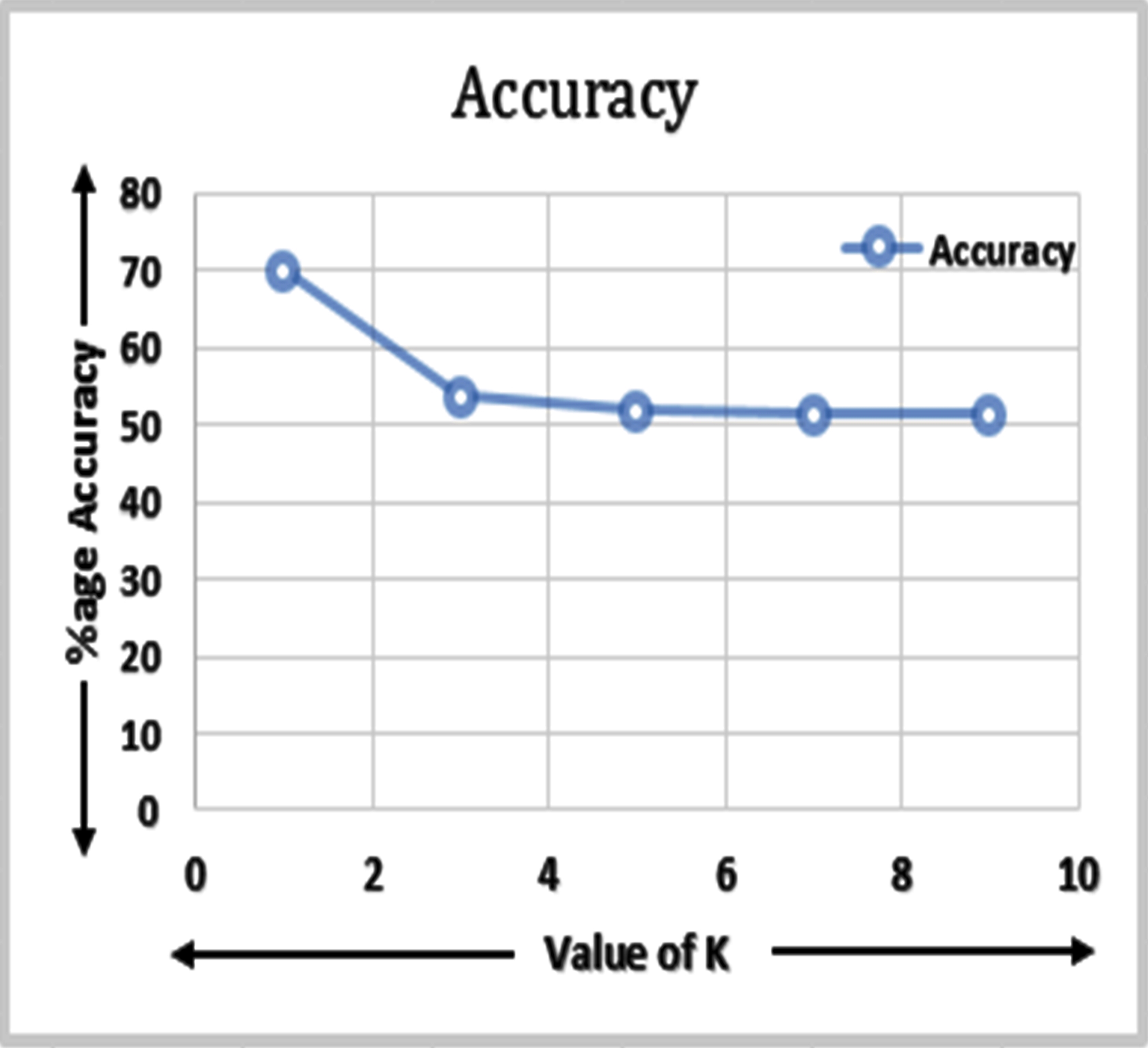}\\
  \caption{KNN accuracy results for different values of K.}\label{fig:KNNgraph}
\end{figure}
\begin{figure}[!t]
  \centering
  \includegraphics[width=2.5in]{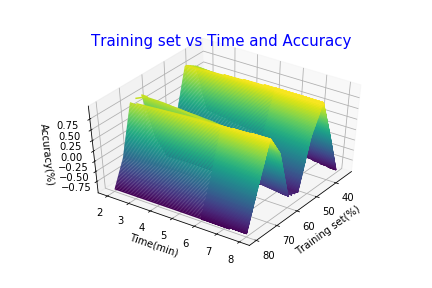}\\
  \caption{NN results of HPLR system.}\label{fig:NNgraph}
\end{figure}
\begin{figure*}[ht]
\centering
\includegraphics[width=5.0in]{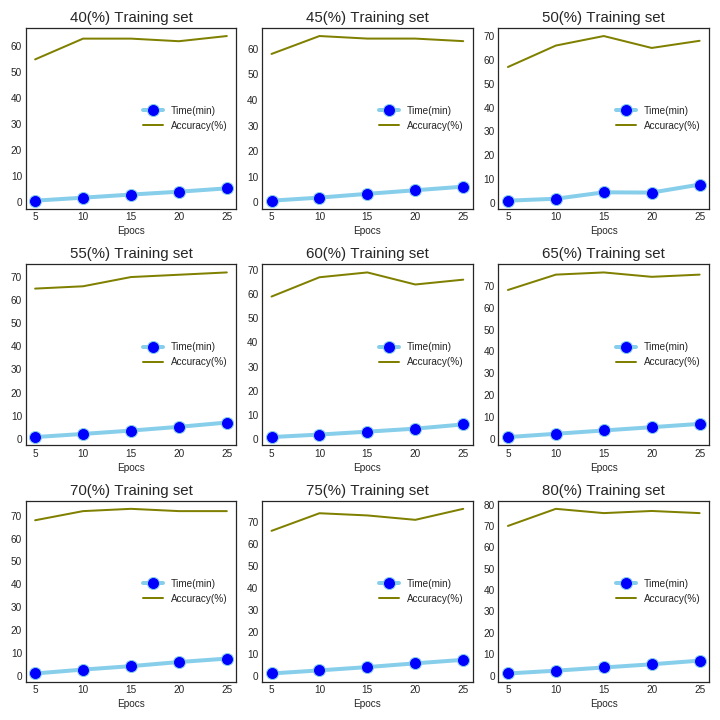}\\
\caption{NN classifier accuracy and time results for varying training and test sets.}\label{fig:NNmulty}
\end{figure*}

\begin{figure}[!t]
  \centering
  \includegraphics[width=2.5in]{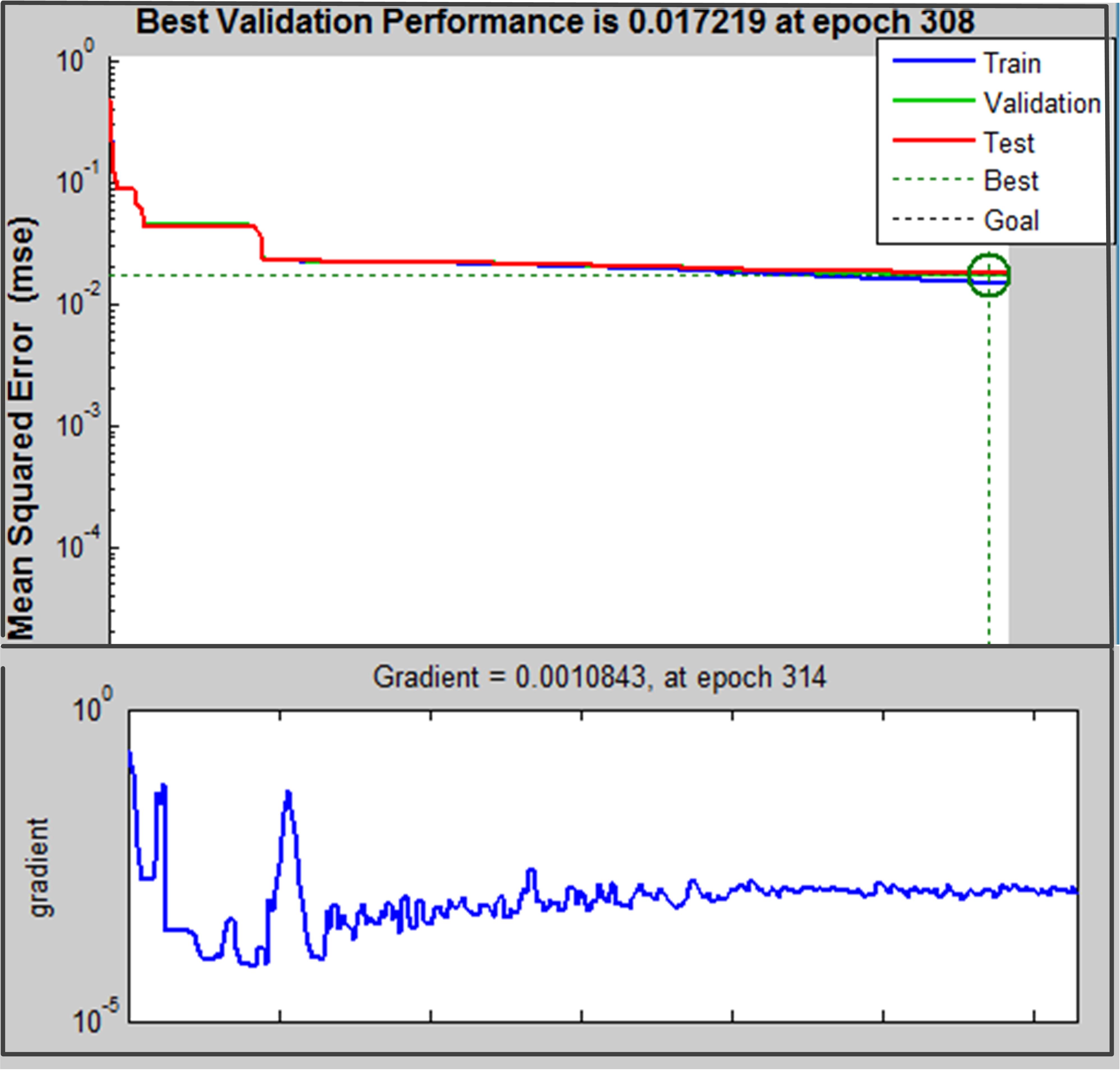}\\
  \caption{MSE and gradient results of HPLR system.}\label{fig:MSE}
\end{figure}

The accuracy of the KNN classifier is tested for different nearest neighbor values of K and it was detected that accuracy varies when the value of K increases, as the occurrence of other class features causes miss-classification. Fig.~\ref{fig:KNNgraph} represents the accuracy results drawn for varying values of K. High accuracy of KNN classifier is noted for the value of K equals to 1, because of high values of K causes the occurrence of other class features that cause miss-classification.

\subsection{Classification Accuracy of Neural Network Classifier}
The feature map is divided into 2:1 for training and test data. NN classifier achieves an accuracy of about 72 \% better than KNN classifier. Fig.~\ref{fig:NNgraph} represents the overall result of NN classifier for Pashto letter recognition problem. 

The efficiency of the classifier is tested for different size of training and test samples vs. time. The data is split into (training, test) sets of of (35\%, 65\%), (40\%, 60\%), (45\%, 55\%), (50\%, 50\%), (55\%, 45\%), (60\%, 40\%), (65\%, 35\%), (70\%, 30\%), (75\%,25\%), and (80\%, 20\%). The corresponding time and accuracy results are generated in Fig.~\ref{fig:NNgraph}. Where it is explained that when there is an increase in the training size, accuracy of the system increases. However, increasing the training size adversely affects the simulation time. A higher accuracy rate of 72\% is carried out for 80\% of training and 20\% of test set.

Furthermore, the NN results based on varying epoch size for different training and test sets are also shown in Fig.~\ref{fig:NNmulty}. It is evident that as the number of epoch increases for given training and test sets, accuracy of the system increases. The mean square error error rate and gradient in the shape for handwritten Pashto letters is shown in Fig.~\ref{fig:MSE}.

\section{Conclusions and Future work}
\label{sec:Conclusions and Future work}
In this paper, an OCR system for automatic recognition of Pashto letters is developed by using KNN and NN classifiers based on zoning feature extractor technique. Experimental results show an accuracy of 70.07\% for KNN while 72\% for NN. Contributions include the provision of handwritten Pashto letters database as a resource for future research work and the experimental results, which will provide a baseline accuracy for future models tested on the data.

In future, we aim to extend and evaluate our technique for a larger database of Pashto script using an increasing number of hidden layers coupled with different feature extractor techniques to achieve a higher accuracy. Furthermore, our goal is to extend the proposed model for the connected letters.

\bibliographystyle{IEEEtran}
\bibliography{references}

\end{document}